\documentclass[a4paper,fleqn]{cas-dc}

\usepackage[numbers,sort&compress]{natbib}
\usepackage{graphicx}
\usepackage{color}
\graphicspath{{}}
\def\tsc#1{\csdef{#1}{\textsc{\lowercase{#1}}\xspace}}
\tsc{WGM}
\tsc{QE}

\begin{document}
\let\WriteBookmarks\relax
\def\floatpagepagefraction{1}
\def\textpagefraction{.001}
\let\printorcid\relax 

\shorttitle{Mesh Deformation-Based Single-View 3D Reconstruction of Thin Eyeglasses Frames with Differentiable Rendering}    

\shortauthors{F.Zhang, Z.Ji, W.Kang et al.}

\title[mode = title]{Mesh Deformation-Based Single-View 3D Reconstruction of Thin Eyeglasses Frames with Differentiable Rendering}

 \cortext[1]{Corresponding author.} 
\author[1]{Fan Zhang}
\ead{13291261757@163.com} 
\author[1]{Ziyue Ji}
\ead{2864235399@qq.com} 
\author[1]{Weiguang Kang}
\ead{kwg0315@163.com} 
\author[2]{Weiqing Li}
\ead{li_weiqing@njust.edu.cn} 
\author[1]{Zhiyong Su}
\cormark[1]
\ead{su@njust.edu.cn} 
\address[1]{School of Automation, Nanjing University of Science and Technology, China}
\address[2]{School of Computer Science and Engineering, Nanjing University of Science and Technology, China}
\begin{abstract}
With the support of Virtual Reality (VR) and Augmented Reality (AR) technologies, the 3D virtual eyeglasses try-on application is well on its way to becoming a new trending solution that offers a "try on" option to select the perfect pair of eyeglasses at the comfort of your own home.
Reconstructing eyeglasses frames from a single image with traditional depth and image-based methods is extremely difficult due to their unique characteristics such as lack of sufficient texture features, thin elements, and severe self-occlusions. 
In this paper, we propose the first mesh deformation-based reconstruction framework for recovering high-precision 3D full-frame eyeglasses models from a single RGB image, leveraging prior and domain-specific knowledge.
Specifically, based on the construction of a synthetic eyeglasses frame dataset, we first define a class-specific eyeglasses frame template with pre-defined keypoints. 
Then, given an input eyeglasses frame image with thin structure and few texture features, we design a keypoint detector and refiner to detect predefined keypoints in a coarse-to-fine manner to estimate the camera pose accurately.
After that, using differentiable rendering, we propose a novel optimization approach for producing correct geometry by progressively performing free-form deformation (FFD) on the template mesh. 
We define a series of loss functions to enforce consistency between the rendered result and the corresponding RGB input, utilizing constraints from inherent structure, silhouettes, keypoints, per-pixel shading information, and so on.
Experimental results on both the synthetic dataset and real images demonstrate the effectiveness of the proposed algorithm.
\end{abstract}



\maketitle

\section{Introduction}
\label{sec:introduction}

With the rapid development of e-commerce, Virtual Reality (VR) and Augmented Reality (AR) technologies have been increasingly used in diverse online shopping domains \cite{Zhangbp18,Marellid21,Marellid22}, allowing people to virtually check the appearance of accessories, clothes, hairstyles, and so on.
Among these, the 3D virtual eyeglasses try-on application is a new trending solution that offers a "try on" option to select the perfect pair of eyeglasses at the comfort of your home.
It enables users to interactively browse and virtually try on eyeglasses to see how they fit, as well as preview their appearance from various viewpoints, improving the senses of reality and immersion \cite{Zhangbp18,FENG2018226,Marellid21}. 
Another advantage is that it can speed up the process of trying several items without having to visit a physical store.

Generally, a pair of eyeglasses is a typical modular product that is made up of an eyeglasses frame and lens based on the functions.
Eyeglasses frames can be divided into three main categories: full-frame, semi-rimless, and rimless glasses. 
They can be made from metal or plastic and come in a variety of shapes and colors.
The eyeglasses frame can further be broken down into different thin elements, such as frame, temple, bridge, tip, nose pads and so on.
In most cases, the width of the thickest part of the frame is usually less than 5 mm. 
These thin elements have insufficient surface detail and few textures to extract features. 
What was worse, some kinds of eyeglasses frames may be uniform in color.
As a result, it is extremely challenging to reconstruct eyeglasses frames with thin structures using existing depth-based or image-based reconstruction methods \cite{2018Liu,2020Vid2Curve,2021root,Pix2Mesh,PMVS,MVS}.

Depth-based reconstruction approaches usually employ RGBD depth sensors to align and integrate depth scans using a truncated signed distance field (TSDF) representation, and then extract a fused surface \cite{Newcombe11,Dai17,2018Liu}.
These methods, such as KinectFusion \cite{Newcombe11} and BundleFusion \cite{Dai17},  have successfully achieved impressive reconstruction results for relatively large structures and environments.
However, due to the low resolution of most depth cameras, they may fail to capture eyeglasses frames primarily consisting of thin or filamentary structures.

Image-based reconstruction methods aim to recover 3D shapes from one or multiple 2D images, which is a long-standing ill-posed problem \cite{2021ImageReview}. 
Recently, 3D reconstruction from a single image using deep learning techniques has been a promising direction and has achieved great success \cite{Pix2Mesh,zhuh21,Maoah21,Zhangjb22,liu2021single}, owing to the superior learning capability to integrate the shape prior.
Taking the advantage of deep neural networks, the output estimated 3D shape is usually represented as a volume \cite{2020Pix2Vox,kim2023volume}, a point cloud \cite{Fanhq17,Insafutdinove18}, or a mesh \cite{Pix2Mesh,Zhangjb22}.
Both volumetric representation and point clouds perform poorly in expressing geometry details and may generate missing parts or broken structures \cite{Maoah21,Pix2Mesh}.
Mesh-based representation has recently gained popularity due to its flexibility and effectiveness in modeling geometric details \cite{Pix2Mesh,Maoah21,Zhangjb22,zhang2023vertex}.

\begin{figure}[!t]
	\centering
	\includegraphics[width=0.5\textwidth]{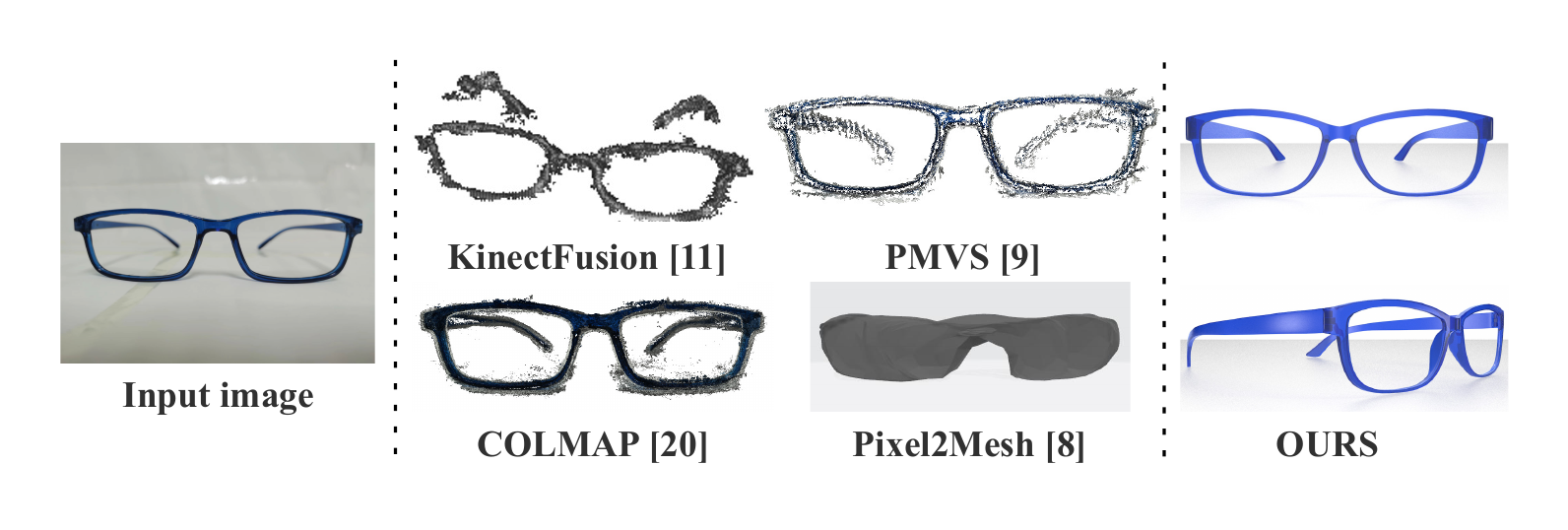}
	\caption{Reconstruction results of different approaches for an eyeglasses frame in the input image.}
	\label{Fig:MethodComparisonIntro}
\end{figure}

Despite the significant success of current cutting-edge techniques, it is still quite challenging for these general frameworks to deal with eyeglasses frames with thin structures. 
On the one hand, RGB images of eyeglasses frames lack sufficient texture features, and are only a few pixels wide in some places.
As a result, they have few textures for feature extraction and matching.
On the other hand, despite the fact that existing general reconstruction methods have significantly improved the reconstruction quality by utilizing refinement modules, they still perform poorly or fail to recover thin and small parts \cite{2021ImageReview}. 
Fig. \ref{Fig:MethodComparisonIntro} shows typical results for an input image using different reconstruction approaches: KinectFusion \cite{Newcombe11}, PMVS \cite{PMVS}, COLMAP \cite{COLMAP}, and Pix2Mesh \cite{Pix2Mesh}. 
It can be seen that the thin parts are either noisy or missing in the fused output.

To address the aforementioned challenges, this paper proposes the first mesh deformation-based reconstruction framework to recover a high-precision 3D full-frame eyeglasses model from a single RGB image.
The key idea is to define a mesh-based eyeglasses frame template and then exploit prior and domain-specific knowledge to deform it into concrete mesh using an elaborate unsupervised free-form deformation (FFD) technique which incorporates the differentiable rendering.
To enable detail-preserving 3D reconstruction while also accelerating convergence, we first define a class-specific mesh template based on a newly constructed eyeglasses frame dataset.
Then, considering that RGB images of thin eyeglasses lack sufficient texture features, we define a series of keypoints on the frame and devise a coarse-to-fine refinement scheme to detect predefined keypoints to estimate the camera parameters of the input image.
After that, we perform the FFD on the template, and employ the differentiable render \cite{nr} to update FFD control-point positions by enforcing the consistency between the rendered template and the corresponding RGB input.
We integrate symmetry loss, projection loss, and regularization loss to allow accurate reconstruction.
Our main contributions are as follows: 
\begin{itemize}
	\item We propose the first mesh deformation-based single-view 3D reconstruction framework for thin full-frame eyeglasses.
	
	\item We establish a synthetic eyeglasses frame mesh dataset with six typical types of eyeglasses frames and nine generic sizes each, with each mesh associated with a series of predefined keypoints. 
	
	\item We introduce a coarse-to-fine network to detect and refine predefined keypoints to accurately compute camera parameters.
	
	\item We develop an unsupervised FFD technique for deforming the class-specific mesh template to iteratively refine the reconstructed mesh via differentiable rendering.

\end{itemize}

\section{Related Work}

In this section, we will first review image-based 3D reconstruction methods based on deep learning for generic objects, followed by a discussion of related works for thin structure reconstruction.

\subsection{Image-based 3D Reconstruction} 

Recently, image-based 3D reconstruction methods using deep learning techniques have led to a new generation of approaches capable of inferring the 3D geometry and structure of objects and scenes from one or more 2D images \cite{image_review}.
According to the representation of the output, these methods can be divided into three categories: volumetric-based, surface-based, and intermediation-based methods. 

\subsubsection{Volumetric-based Methods}

Volumetric-based methods generally discritize the space around 3D objects into regular voxel grids.
These methods are usually very expensive in terms of computation and memory requirements.
And, volumetric models are very sparse since surface elements are contained in a few voxels. 
In the literature, there are four main volumetric representations: binary occupancy grid, probabilistic occupancy grid, Signed Distance Function (SDF), and truncated signed distance function (TSDF).
The Binary occupancy grid is based on the assumption that each voxel is either occupied or unoccupied \cite{20163D-R2N2,2017MVS,2019Pix2Vox,20213D-VRVT}.
Instead of a binary state, each voxel in the probabilistic occupancy grid represents its probability of belonging to the objects of interest \cite{20183D-RCNN,2018LHLV,2021HQ}. 
For the SDF, each voxel encodes its signed orthogonal distance of a given point to the boundary of a subset of a metric space, such as Euclidean space \cite{2019DeepSDF}.
The sign is negative if the voxel is inside the object. 
Otherwise, the sign is positive. 
The TSDF estimates distances along a range sensor's lines of sight to form a projective signed distance field, which is then truncated at small negative and positive values \cite{2022NeuralRGBD,2017MarrNet}.

\subsubsection{Surface-based Methods}

Surface-based methods explore representations such as meshes and point clouds, since information is rich only on or near the surfaces of 3D shapes.
These approaches can be classified into three main categories: parameterization-based,  deformation-based, and point-based methods \cite{2021ImageReview}. 
Parameterization-based methods represent the surface of a 3D shape as a mapping from a regular parameterization domain to 3D space.
They are suitable for the reconstruction of objects of a specific shape category, such as human faces and bodies \cite{3dmm,SCAPE,SMPL,2015Dyna,2019Expressive}.
These methods, however, can only be used to recover shapes with fixed topology.
Deformation-based methods estimate a deformation field from an input image, which is then applied to a template 3D shape to produce the reconstructed 3D model \cite{nr,Pix2Mesh,2018FFD}.
Point-based techniques employ an unordered set of points to represent a 3D shape, which is well suited for objects with intriguing parts and fine details \cite{2018Multiresolution,2018GAL,2022Point-NeRF}.
They are capable of handling 3D shapes with arbitrary topologies. 
However, they require a post-processing procedure to recover the 3D surface mesh.

\begin{figure*}[!t]
	\centering
	\includegraphics[width=1\textwidth]{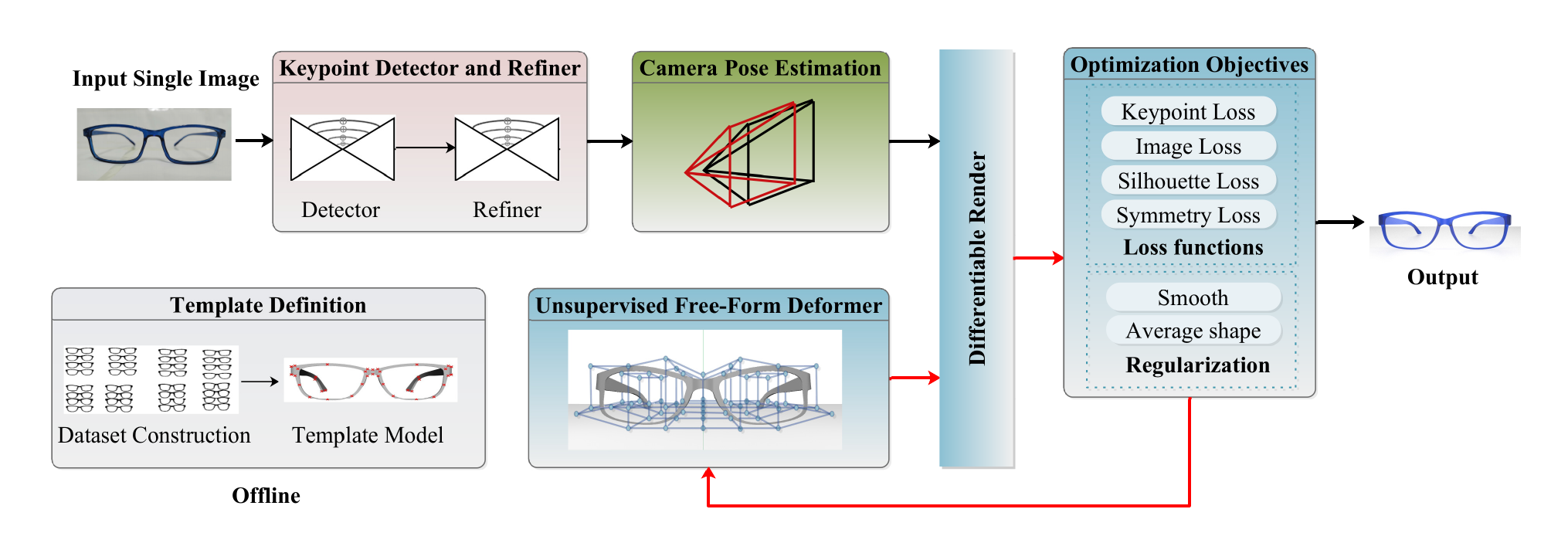}
	\caption{ Overview of the proposed reconstruction framework. During the offline phase, we first build a 3D eyeglasses frame model dataset, and then define a template mesh associated with 42 keypoints. In the online phase, given an input RGB image, after estimating the keypoints and camera pose, the unsupervised free-form deformer progressively deforms the template mesh to enforce consistency between the rendered result and the input RGB image through the differentiable rendering in an iterative manner. The red arrow part represents the progress of our online optimization iteration.}
	\label{Fig:Overview}
\end{figure*}

\subsubsection{Intermediation-based Methods}

Instead of directly predicting an object's 3D geometry from RGB images, intermediation-based methods decompose the problem into sequential steps.
And, for each step, an intermediate representation is predicted, such as depth maps, normal maps, and segmentation masks \cite{2018Pix3D,2018LearingPriors}.
The main advantage of these approaches is that the intermediate representations are much easier to recover from 2D images. 
Consequently, recovering 3D models from these three modalities is much easier than from 2D images alone.

\subsection{Thin Structure Reconstruction}
\label{subsec:thinReconstruction}

Despite their impressive results on textured surfaces, existing reconstruction approaches are fragile when dealing with thin structures that lack sufficient surface details.
In this section, we will provide a brief review on methods that focus on thin structure reconstruction from the following two aspects: depth-based, and image-based methods.

\subsubsection{Depth-based Thin Structure Reconstruction}

Depth-based methods use depth sensors to actively measure depth for thin structure reconstruction.
Li et al. present a novel deformable model, called the arterial snake, to reconstruct objects with delicate structures, which are created using cane, coils, metal wires, rods, etc \cite{Li2010Snake}.
Liu et al. propose the first approach, named CurveFusion, for high quality scanning of thin structures using a handheld RGBD camera \cite{2018Liu}.
Their method is only applicable in certain scene conditions, such as indoor environments without strong sunlight and objects with non-black surface colors. 
However, even with advanced acquisition technologies, scanning thin structures has proven to be a challenging task, owing to the limited sensor resolution and noisy measured depth \cite{2020Vid2Curve}.

\subsubsection{Image-Based Thin Structure Reconstruction}

Image-based approaches use a single or multiple images to reconstruct thin structures.  
Lu et al. propose an unsupervised learning scheme to reconstruct thin plant root systems from a single image \cite{2021root}.
They first incorporate a cross-view GAN-based network into the reconstruction process to predict the root image from a different perspective in order to reduce reconstruction errors. 
The root system is then rebuilt using stereo reconstruction based on the input image and synthetic views. 
Tabb et al. reconstruct thin, texture-less objects from multiple images using silhouette probability maps by formulating the difference between input and reconstruction images as a pseudo-Boolean minimization problem in a volumetric representation \cite{2013Tabb}. 
Hsiao et al. present a computational framework to reconstruct wire art models from several views of 2D line-drawings \cite{Hsiao18}.
They use constrained 3D path finding in a volumetric grid to resolve spatial ambiguities due to the inconsistency between input line drawings. 
Liu et al. reconstruct continuous 3D wire models from a small number of images using a candidate selection strategy for image correspondence, where each wire is composed of an ordered set of 3D curve segments \cite{Liulj17}.
Li et al. present a novel thin surface reconstruction method using both point clouds and spatial curves generated from image edges \cite{Lisw18}. 
Wang et al. propose the first approach that simultaneously estimates camera motion and reconstructs the geometry of complex 3D thin structures in high quality from a color video captured by a handheld camera \cite{2020Vid2Curve}. 
These methods mainly aim to reconstruct 1D thin structures with circular sections (e.g., pylons and wires), which cannot be extended to generic thin objects.

All in all, to the best of our knowledge, no reconstruction algorithm has been specially designed for thin eyeglasses frames in the literature.
Existing reconstruction methods discussed above are all designed for specific categories.
These methods exploit prior knowledge of specific objects with thin structures for the reconstruction, which are not suitable for the reconstruction of thin eyeglasses frames.

\begin{figure*}[!htbp]
	\centering
	\includegraphics[width=1.0\textwidth]{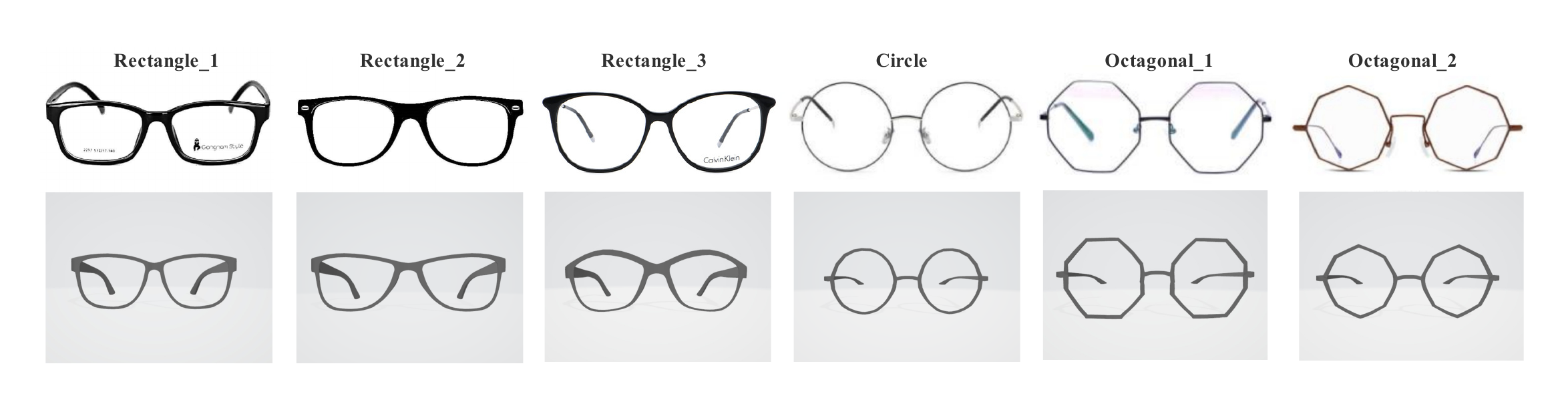}
	\caption{Examples of six typical kinds of eyeglasses frames (e.g., rectangle, octagon, and circle) and their 3D models in our dataset.}
	\label{Fig:dataset}
\end{figure*}
\section{Methods}

\subsection{Overview}

We propose a mesh deformation-based reconstruction framework to recover a high-precision 3D full-frame eyeglasses model from a single RGB image, which is characterized by thin structure and insufficient texture features.
The entire framework consists of template definition, keypoint detector and refiner, camera pose estimator, and unsupervised free-form deformer, as illustrated in Fig.\ref{Fig:Overview}.
In the offline stage, the template definition computes an appropriate shape as a class-specific mesh template from the constructed 3D eyeglasses frame model dataset and defines 42 keypoints on the template. 
In the online stage, firstly, the keypoint detector and refiner takes a single-view RGB image of an eyeglasses frame as input and outputs detected keypoints in the image.
Then, the camera pose estimator estimates the camera pose by minimizing the distance between the projection of 3D keypoints of the template and detected 2D keypoints. 
After that, the unsupervised free-form deformer defines a FFD grid and estimates a symmetric FFD displacement to progressively deform the template mesh into the desired 3D model. 
We adopt a differentiable render to render the deformed template mesh to an image and design a series of optimization objectives to minimize the distance between the input image and the rendered image in an iterative way. 

\subsection{Template Definition}

To achieve high fidelity reconstruction results, we resort to defining a full-frame eyeglasses template for subsequent reconstruction.
To this end, we first construct a dataset of 3D eyeglasses frame models.
Then, we use the built dataset to find an appropriate shape to use as a class-specific mesh template.
Finally, to facilitate the following keypoint detection, we define 42 keypoints on the template according to structural characteristics.

\subsubsection{Dataset Construction}

In this paper, we employ the Cinema 4D \cite{mcquilkin2011cinema} modeling software to create a total of 54 3D full-frame eyeglass models in our dataset $\mathbb{D}$, which can be classified into three categories, including rectangle style, octagon style, and circle style, as shown in Fig.\ref{Fig:dataset}. 
We start with an initial eyeglass frame model and use the Mesh Deformer Tool of Cinema 4D to embed it into a control grid. 
Then, we move the control points on the grid to deform the initial model to generate various models with different shapes and sizes.
Finally, our dataset consists of six different types of eyeglass frame models, as illustrated in Fig.\ref{Fig:dataset}.
Each type of frame comes in nine different sizes that adhere to the China national standard GB/T 14214. 
Each eyeglass frame model has 13768 vertices and 15664 faces.
All the eyeglass frame models in our dataset are aligned and then are used for the following template definition.

\subsubsection{Template Calculation}

The template calculation aims to compute an appropriate template model from previously constructed dataset in the offline stage to speed up the convergence during the deforming process and allow detail-preserving 3D reconstruction. 
For a given dataset $\mathbb{D}$ with $n_{m}$ eyeglasses frame models, this problem can be expressed mathematically as
\begin{flalign}
&&
\label{eq:d}
\underset{M}{\arg \min}  \frac{1}{n_{m}}\sum_{i=1}^{n_{m}}Dist(M, D_{i}),
&&
\end{flalign}
where $M$ represents the desired template model, $D_{i}$ represents the $i$-th model in the dataset $\mathbb{D}$.
The $Dist(M, D_{i})$ function calculates the Euclidean distance between the vertices of $M$ and $D_{i}$.
Eq.(\ref{eq:d}) is a typical Fermat-Torricelli problem \cite{2002GM}. 

The Weiszfeld algorithm \cite{2016Weiszfeld}, which is a form of iteratively re-weighted least squares, is employed to calculate the desired template model $M$.
It defines a set of weights that are inversely proportional to the distances from the current estimate to the sample model and creates a new estimation that is the weighted average of the sample according to these weights.
Firstly, we calculate the arithmetic mean model $M_{0}$ of all models as the initial model of $M$:
\begin{flalign}
&&
    M_{0} = \frac{1}{n_{m}}\sum_{i=1}^{n_{m}}D_{i}.
    &&
\end{flalign}
Then, we use the following iteration formula to update the current model $M_{i+1}$:
\begin{flalign}
&&
    M_{i+1} = (\sum_{i=1}^{n_{m}}\frac{D_{i}}{Dist(M_{i}, D_{i})})/(\sum_{i=1}^{n_{m}}\frac{1}{Dist(M_{i}, D_{i})}).
    &&
\end{flalign}


Finally, we can get the desired template model $M = (\mathcal{V}, \mathcal{F})$ after about 40 iterations, which is represented with vertices $\mathcal{V} = \{ v_{1}, \cdots, v_{n_{v}} \}$ and faces $\mathcal{F}$.
In this paper, $n_{v} = |\mathcal{V}| = 13768$, and $|\mathcal{F}| = 15664$.
The defined template model $M$ will be used in the subsequent camera pose estimation and unsupervised free-form deformer, as illustrated in Fig.\ref{Fig:Keypoints}.

\subsubsection{Keypoint Definition}

\begin{figure}[!t]
	\centering
	\includegraphics[width=0.45\textwidth]{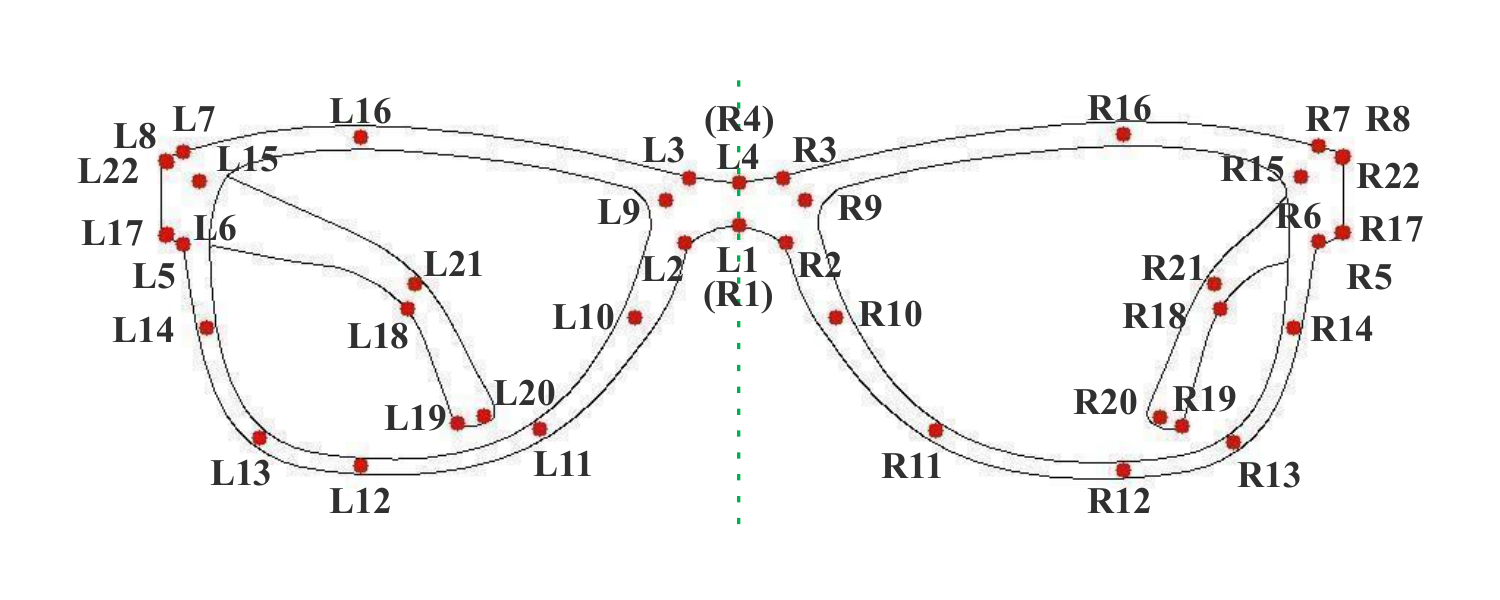}
	\caption{Predefined keypoints on the template.}
	\label{Fig:Keypoints}
\end{figure}

The keypoint definition targets to define a series of keypoints on the 3D template mesh to facilitate the following camera pose estimation as well as template deformation, since RGB images of thin eyeglasses always lack sufficient texture features.
These 3D keypoints are required to be able to reflect the essential characteristics of a shape and to establish reliable shape representation.
We select a total of 42 keypoints on the 3D template frame manually and record their vertex indexes, as shown in Fig.\ref{Fig:Keypoints}.
Among the predefined keypoints, the front gets 30 keypoints, which determine the style and width of the frame. 
The temple has 12 keypoints, which determine the length and pantoscopic angle. 
By this way, the keypoints of each frame in the dataset $\mathbb{D}$ can be easily retrieved through their vertex indexes, since all frame models are aligned.

To train the following keypoint detection and refinement module, for each frame mode $D_{i}$ in $\mathbb{D}$, we also render 845 images associated with corresponding 2D keypoints from different views by using the OpenSceneGraph \cite{burns2004open} software.
Thus, there are $54 \times 845 = 45630$ image-keypoint pairs for training in total.  

\begin{figure*}[!htbp]
	\centering
	\includegraphics[width=1.0\textwidth]{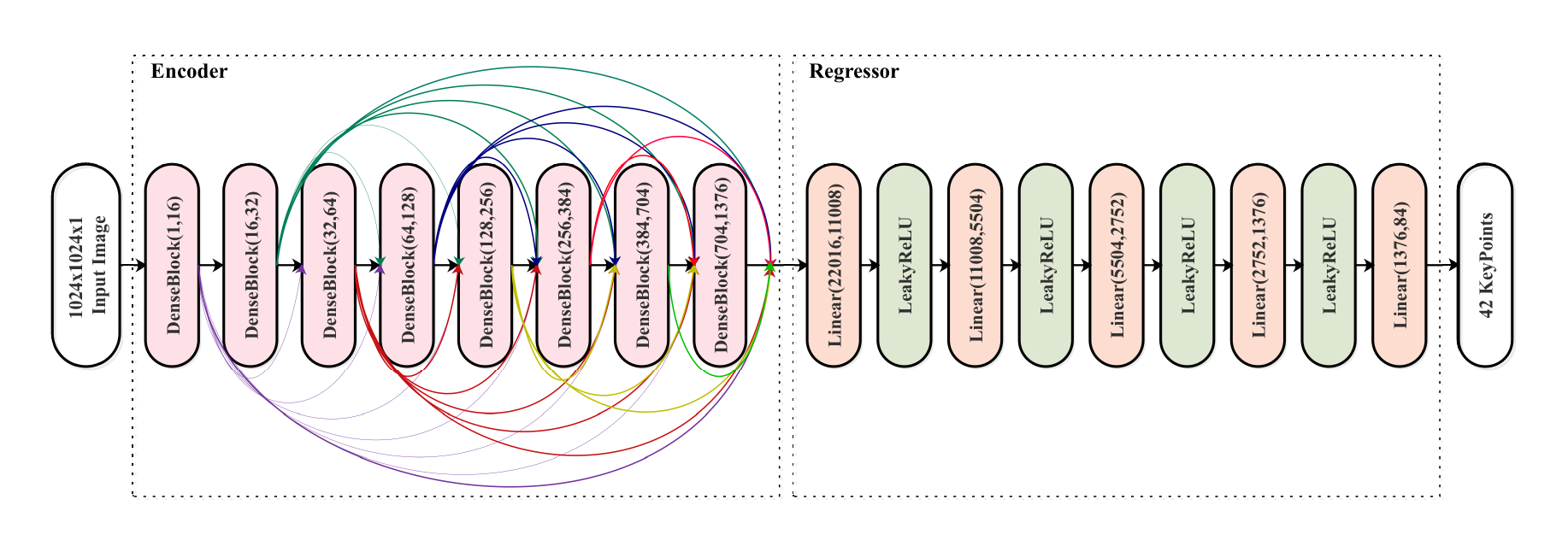}
	\caption{Network architecture of the keypoint detector. The input of a dense block comes from every output of the foregoing dense block. The output of the encoder is squeezed into a vector and is regressed into keypoint coordinates by an MLP.}
	\label{Fig:KeypointDetector}
\end{figure*}

\subsection{Keypoint Detector and Refiner}

Given a single-view RGB image of an eyeglasses frame as input, the keypoint detector and refiner adopt a coarse-to-fine strategy to detect accurate predefined keypoints, which are then used for the following camera pose estimation and FFD.

\subsubsection{Keypoint Detector}

The keypoint detector, which consists of an encoder and a regressor, aims to detect predefined 2D keypoints from the input single-view RGB image, as illustrated in Fig.\ref{Fig:KeypointDetector}. 

The keypoint detector uses a Mean Square Error (MSE) as the loss function,
\begin{flalign}
&&
\label{eq:kp2d}
	L_{kp2d}=\frac{1}{42} \sum_{i=1}^{42} \left \| p^{d}_{i} - p^{g}_{i} \right \|,
 &&
\end{flalign} 
where $p^{d}_{i}$ is the $i$-th 2D keypoint coordinate given by the detector, and $p^{g}_{i}$ is the ground-truth coordinate of the corresponding 2D keypoint. 
Note that it is difficult to directly regress accurate keypoint coordinates from the input image when it lacks sufficient texture information. 
Therefore, the 2D keypoints detected by the keypoint detector inevitably suffer from offsets. 

\begin{figure*}[!htbp]
\centering
\includegraphics[width=0.7\textwidth]{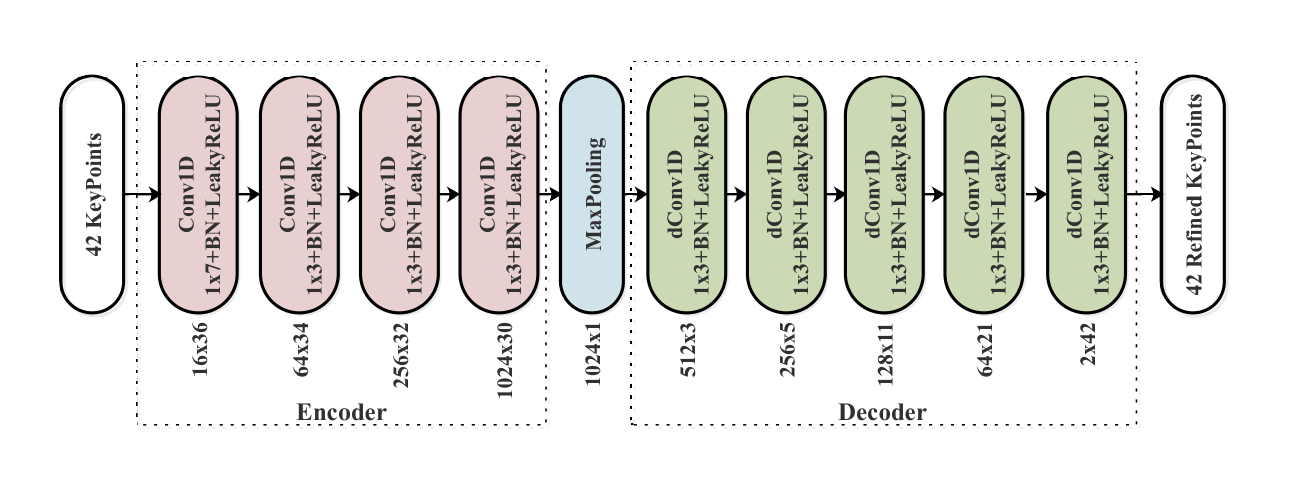}
\caption{Network architecture of the keypoint refiner. The convolution layers and maxpooling layer encode the rough coordinates of 2D keypoints into a feature embedding with a  dimension of $1024 \times 1$. Afterwards, the decoder decodes the feature embedding into more accurate coordinates of 2D keypoints.}
\label{Fig:Refiner}
\end{figure*}

\subsubsection{Keypoint Refiner}

To improve the reconstruction quality, the keypoint refiner attempts to predict refined coordinates $p^{r}$ with higher accuracy from the keypoint detector's coarse estimation $p^{d}$. 
The refiner, which uses a denoising encoder-decoder architecture to reduce high-frequency offset, contains three parts: an encoder, a maxpooling layer, and a decoder, as depicted in Fig. \ref{Fig:Refiner}.

\subsection{Camera Pose Estimation}

The camera pose is represented by the camera position $c^{p} \in \mathbb{R}^{3}$ and direction $c^{d} \in \mathbb{R}^{3}$.
Our algorithm finds the optimal camera pose by minimizing the reprojection error, given as:
\begin{flalign} 
&&
\label{eq:l-camerapose-argmin}
\arg\min_{c^{p}, c^{d}} \sum_{i=1}^{42} \| Proj(v_{i}^{k}, c^{p}, c^{d})-p_{i}^{r} \|^2,
&&
\end{flalign}
where $v_{i}^{k}$ is the $i$-th predefined 3D keypoint, $p_{i}^{r}$ is $i$-th refined 2D keypoint in the input image, and $Proj(v_{i}^{k},c^{p}, c^{d})$ is a projection function that projects the 3D keypoint $v_{i}^{k}$ on the template model to the image space according to $c^{p}$ and $c^{d}$.
We use the distance between the projected 2D keypoints obtained by $Proj(v_{i}^{k}, c^{p}, c^{d})$ and the corresponding refined 2D keypoints $p_{i}^{r}$ as the loss function to optimize the six parameters of $c^{p}$ and $c^{d}$.
Once the optimal camera pose is obtained, it will be fixed during the deformation process.

\subsection{Unsupervised Free-Form Deformer}

The proposed unsupervised free-form deformer estimates a deformation field $\Delta$, which, when applied to the template mesh $M$, results in the reconstructed 3D eyeglasses frame model $\tilde{M}$.
The motivation for using FFD is to fully leverage the prior knowledge provided by the template eyeglasses model. 
It ensures that the relative positions of vertices in the reconstructed eyeglass frame model remain stable during the deformation process, thereby preserving the reasonable geometric structure of the eyeglass frames and resulting in smoother reconstruction.
To that end, we define various structure and image-related losses to capture properties of various aspects and employ the differentiable rendering \cite{nr} to enforce consistency between the rendered result and the corresponding RGB input.

\subsubsection{Free-Form Deformation}

\begin{figure}[!t]
\centering
	\includegraphics[width=0.5\textwidth]{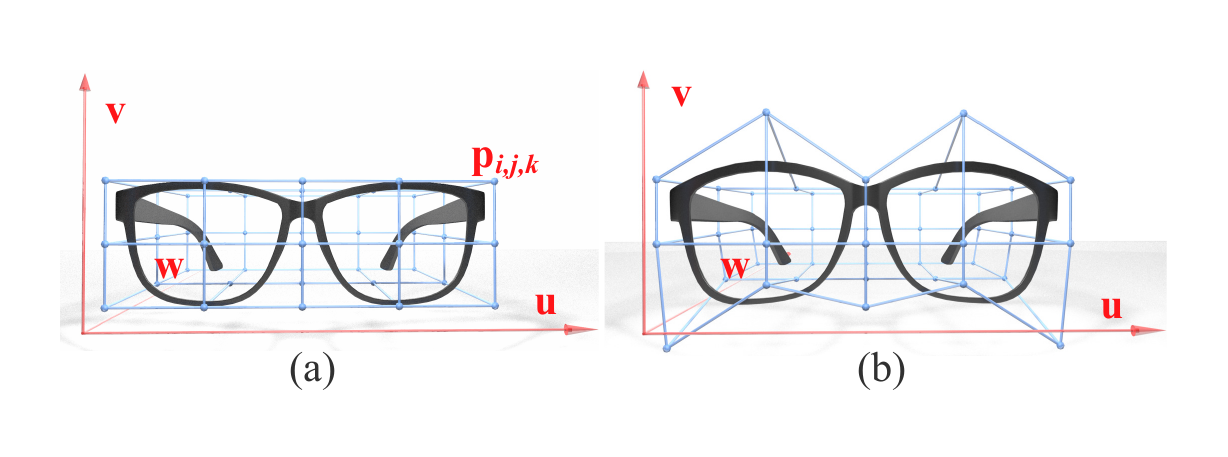}
	\caption{Illustration of the FFD lattice of control points with $l = 4$, $m = 2$, $n = 2$. (a) The local coordinate system $(u, v, w)$ and the control points $p_{i,j,k}$ ; (b) The template is deformed by simply free manipulating control points.}
	\label{Fig:FFD}
\end{figure}

The FFD technique, which is a three-dimensional extension of the Bezier curve form and has been widely used in shape deformation\cite{Sederberg86,Kaesemodel17,Kurenkova18,Jackd19}, is used to gradually deform the eyeglasses frame template $M$ by freely manipulating control points.
The fundamental idea behind FFD is that it deforms the space around the template $M$ rather than directly deforming its vertices, as illustrated in Fig.\ref{Fig:FFD}. 
Specifically, the FFD creates a parallelepiped-shaped lattice of $n_{c}$ control points with axes defined by the orthogonal vectors $\mathbf{u}$, $\mathbf{v}$ and $\mathbf{w}$ \cite{Sederberg86}, where $n_{c} = l \times m \times n $.
The control points are defined by $l$, $m$ and $n$ which divide the lattice in $l+1$, $m+1$, $n+1$ planes in the $\mathbf{u}$, $\mathbf{v}$, $\mathbf{w}$ directions, respectively.
Therefore, any vertex in the lattice space can be represented as a linear combination of control points.
And, the shape of the template $M$ can be deformed indirectly by manipulating control points in the lattice space. 

Let $v_{ffd} = (u, v, w)$ be the vertex of the template $M$ in the lattice space, and $\delta_{i,j,k}$ be the 3D deformation offset at the $i$, $j$, $k$-th control point $p_{i,j,k} = (i,j,k)$. 
Then, the deformed position $\tilde{v}_{ffd}$ of any arbitrary vertex $v_{ffd}$ in the lattice space is given by
\begin{flalign}
&&
\label{eq:vffd}
	\begin{aligned}
	&\tilde{v}_{ffd}(u, v, w) = \\
	&\sum_{i=0}^{l} \sum_{j=0}^{m} \sum_{k=0}^{n} \left(p_{i,j,k} + \delta_{i,j,k}\right) B_{il}(u) B_{jm}(v) B_{kn}(w),
	\end{aligned}
 &&
\end{flalign}
where $B_{\theta n} (x)$ is the Bernstein polynomial of degree $n$ which sets the influence of each control point on every vertex of the template mesh $M$ \cite{Kurenkova18}.
Eq. (\ref{eq:vffd}) can also be written in a matrix form as,
\begin{flalign}
&&
\label{eq:mffd}
\tilde{\mathcal{V}}_{ffd} = \mathbf{B} ( \mathcal{P} + \Delta),
&&
\end{flalign}
where $\tilde{\mathcal{V}}_{ffd} \in \mathbb{R}^{n_{v} \times 3}$ are the vertices of the deformed mesh $\tilde{M}$, the deformation matrix $\mathbf{B} \in \mathbb{R}^{n_{v} \times n_{c}}$ is a set of Bernstein polynomial basis \cite{Sederberg86},  $\mathcal{P} \in \mathbb{R}^{n_{c} \times 3}$ are the control point coordinates, $\Delta = \{ \delta_{i,j,k} \} \in \mathbb{R}^{n_{c} \times 3}$ is the deformation field to be optimized, $n_{v}$ and $n_{c}$ are the number of vertices and control points, respectively.
When the deformation field $\Delta$ is applied to control points $\mathcal{P}$, the FFD deforms the entire space around the template mesh $M$ to get the reconstructed mesh $\tilde{M}$.

\subsubsection{Optimization Objectives}

We define a series of objectives to optimize and constrain the FFD procedure to guarantee appealing results.
These objectives can be classified into three categories: symmetry loss $L_{sym}$, projection loss $L_{proj}$, and regularization loss $L_{regu}$.
During optimization, the deformed model with an associated camera pose can be formulated into
a 2D color image $I^{re}$ through a neural renderer.
The ideal result is for the rendered image $I^{re}$ to match the input image $I$. 
To this end, we employ the projection loss $L_{proj}$ to measure the pixel-wise difference.
Besides, to make the results plausible, the regularization loss $L_{regu}$ is also introduced.
Thus, the total optimization objectives for the FFD are defined as follows:
\begin{flalign}
&&
\label{eq:loss}
	L_{opt} = \omega_{sym}L_{sym} + L_{proj} + L_{regu},
 &&
\end{flalign}
where $\omega_{sym}$ is the loss weight.

\textbf{Symmetry Loss $L_{sym}$}. 
Symmetry is an inherent structural feature of eyeglasses frames, as illustrated in Fig. \ref{Fig:Keypoints}.
Exploiting this domain knowledge as a prior, the symmetry loss imposes a vertex-wise symmetric constraint on the deformed mesh, which may effectively alleviate the self-occlusion and viewing direction problems, greatly improving deformation and reconstruction precision.
To that end, in this paper, the template's predefined 3D keypoint set $V$ is first divided into left ($V^{l}$) and right ($V^{r}$) parts based on the plane of symmetry $X = \sigma$, as determined by
\begin{flalign}
&&
	\label{eq:axis-symmetry}
	X = \sigma = \frac{1}{42}\sum_{i=1}^{42} v_{i}(x)/2, 
 &&
\end{flalign}
where $v_{i} \in  V$, $v_{i}(x) $ is $x$-coordinate of the $i$-th keypoint. 
The height ($y$) and depth ($z$) of the 3D keypoints with a symmetrical relationship should be equal. 
Therefore, the symmetry loss takes the following form,
\begin{flalign}
&&
	\label{eq:l-sym}
	\begin{split}
		L_{sym} =  \sum_{i=1}^{21} &(|v^{l}_{i}(x)|+|v^{r}_{i}(x)| - 2 \sigma)^2 +\\
		&(v^{l}_{i}(y) - v^{r}_{i}(y))^2 + (v^{l}_{i}(z) - v^{r}_{i}(z))^2
	\end{split},
 &&
\end{flalign}
where $v^{l}_{i} \in V^{l}$, $v^{r}_{i} \in V^{r}$, $v^{l}_{i}$ and $v^{r}_{i}$ are symmetric keypoints.

\textbf{Projection Loss $L_{proj}$}.
The projection loss consists of three items: image loss $L_{img}$, silhouette loss $L_{sil}$, and keypoint loss $L_{kp}$,
\begin{flalign}
&&
	\label{eq:loss-proj}
	L_{proj} = \omega_{img} L_{img} + \omega_{sil} L_{sil} + \omega_{kp} L_{kp},
 &&
\end{flalign}
where $\omega_{img}$,  $\omega_{sil}$, and $\omega_{kp}$ are the loss weights, respectively.

\begin{itemize}
	\item \textbf{Image Loss $L_{img}$}.
	The image loss $L_{img}$, which is actually a pixel-to-pixel loss, computes the L2 distance for all pixels between the rendered image $I^{re}$ and the input image $I$ to measure the pixel-wise difference,
	\begin{flalign}
 &&
		\label{eq:l-img}
		L_{img} = \sum \left \| I^{re} - I\right \|.
  &&
	\end{flalign}

	\item \textbf{Silhouette Loss $L_{sil}$}.
	The silhouette loss measures the silhouette consistency between the rendered image $I^{re}$ and the input image $I$ based on an Intersection-over-Union (IoU) loss,
	\begin{flalign}
 &&
		L_{sil}=\sum \frac{I^{re} \odot I} {I^{re} \times I}.
		\label{eq:l-sil}
  &&
	\end{flalign}

	\item \textbf{Keypoint Loss $L_{kp}$}.
	The keypoint loss provides position supervision information for 3D keypoints, forcing them to move to the desired location,
	\begin{flalign}
 &&
		\label{eq:l-kp}
		L_{kp}=\frac{1}{42} \sum_{i=1}^{42} \left\| p^{t}_{i} - p^{r}_{i} \right \|,
  &&
	\end{flalign} 
	where $p^{t}_{i}$ is the projection of the $i$-th 3D keypoint of the deformed template model in the image space, $p^{r}_{i}$ is the refined detected 2D keypoint of the $i$-th 3D keypoint.
\end{itemize}

\textbf{Regularization Loss $L_{regu}$}.
Even with the symmetry and projection loss mentioned above, the optimization can easily become stuck in some local minimum, such as generating large deformation to favor some local consistency. 
Therefore, two regularization items are designed to make the result more reasonable, including smooth regularization $L_{sm}$ and average shape regularization $L_{avg}$.
\begin{flalign}
&&
	\label{eq:loss-regu}
	L_{regu} = \omega_{sm} L_{sm} + \omega_{avg} L_{avg},
 &&
\end{flalign}
where $\omega_{sm}$,  $\omega_{avg}$ are the loss weights, respectively.

\begin{itemize}
	\item \textbf{Smooth regularization $L_{sm}$}. 
	The smooth regularization $L_{sm}$ adopts the classical Laplacian term \cite{Huber1992} to reduce the sharp places in the result. 
	The Laplace operator weights each vertex  $v_{i}$ and its neighboring vertices $v_{ij}$ to achieve smooth effect,
	\begin{flalign}
 &&
		\label{eq:l-smooth}
		L_{sm}=\frac{1}{n_{v}} \sum_{i=1}^{n_{v}} \left (  v_{i}-\frac{1}{k}\sum_{j=1}^{k}v_{ij}  \right ) ,
  &&
	\end{flalign} 
	where $n_{v}$ is the number of vertices of the template, $k$ is the number of neighbors of $v_{i}$ and is set to 6 in this paper. 
	
	\item \textbf{Average shape regularization $L_{avg}$}.
	Some keypoints occluded in the input image may become unconstrained during deformation, resulting in unpredictable reconstruction results.
	The average shape regularization $L_{avg}$ is designed to achieve the optimal reconstruction result within the minimum deformation, which is defined as
	\begin{flalign}
 &&
		\label{eq:l-avg}
		L_{avg} = \left \| \tilde{\mathcal{V}} - \mathcal{V}\right \|,
  &&
	\end{flalign} 
	where $\mathcal{V}$ and $\tilde{\mathcal{V}}$ are the vertices of the model before and after the deformation.   
 
\end{itemize}

\begin{figure}[!t]
	\centering
	\includegraphics[width=0.25\textwidth]{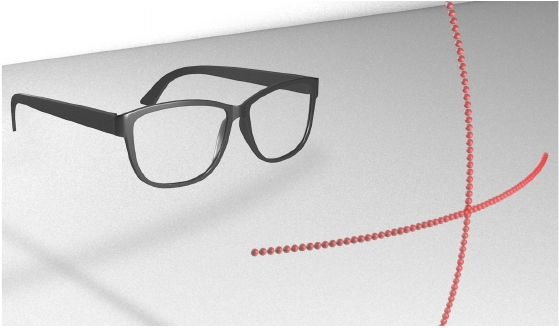}
	\caption{Generation of projected images from the synthetic dataset.}
	\label{Fig:viewpoint}
\end{figure}

\begin{table*}[!t]
	\centering 
	\caption{Details about the FLOPS, memory usage, training time, and the number of optimizable parameters of each module.}
        \resizebox{\linewidth}{!}{
	\begin{tabular}{cccccc}
		\hline    
		Module & FLOPS & Memory usage & Training time & Inference time & Optimizable parameters\\
		\hline
		Keypoint Detector &2.75GMac &2756.90MB &5h &0.25s &323.64M \\
		Keypoint Refiner	& 33.77MMac & 1291.98MB & 2.5h & 0.003s & 2.94M \\
		Camera Pose Estimation 	& - & 1938.40MB & - & 4min & 6 \\
		Unsupervised free-form deformer		& - & 2154.59MB & - & 7min & 1200 \\
		\hline
	\end{tabular}
        }
	\label{Tab:Parameters}
\end{table*}
\section{Experiments}

\subsection{Experimental Setup}

We implement our method with Pytorch \cite{2019Pytorch} on a Windows 10 PC with an Intel Core i7-9700KF processor (3.6 GHz), 16 GB RAM, and a 8 GB NVIDIA GeForce RTX 2070 graphics card.
We quantitatively evaluate the proposed method on our synthetic dataset $\mathbb{D}$,  which contains six types of eyeglasses frames with nine different sizes.
As illustrated in Fig. \ref{Fig:viewpoint}, the camera is oriented towards the center of the eyeglass frame.
The camera position varies within the intervals of [$-30^\circ$, $30^\circ$] around the y-axis and [$-30^\circ$, $30^\circ$] around the x-axis. 
The camera's own rotation angle varies within the interval 
[$-15^\circ$, $13^\circ$], with the rotation axis being the z-axis of the camera's local coordinate system. 
Each image is generated with a step size of $5^\circ$ for camera position and $7^\circ$ for rotation angle. 
Consequently, we generated a total of 
$13 \times 13 \times 5 $ images, each with a resolution of $1024 \times 1024$ and corresponding 2D keypoints for each glass.

During our keypoint detection, we randomly selected 80\% (676) of the images for training and 20\% (169) for testing.
Thus, in total, there are $6 \times 9 \times 676 $ generated images for training, and $6 \times 9 \times 169 $ images for testing.
The SGD optimizer with a learning rate of $5\times10^{-3}$ is used for training the keypoint detector and refiner,
and the network's learning rate will decrease at an exponential rate of 0.92 during training. 
The Adam optimizer with a learning rate of $1 \times 10^{-3}$ is used for the camera pose estimation and the unsupervised free-form deformer.
The loss weights in our free-form deformation $\omega_{sym} = 0.1$, $\omega_{img} = 5 \times 10^{-5}$, $\omega_{sil} = 5 \times 10^{-5}$, $\omega_{kp} = 1$, $\omega_{sm} = 0.012$, and $\omega_{avg} = 2.1 \times 10^{-4}$.
Table \ref{Tab:Parameters} shows details about the FLOPS, memory usage, training time, and the number of optimizable parameters of each module.

\begin{figure*}[!ht]
    \centering
	\includegraphics[width=0.9\textwidth]{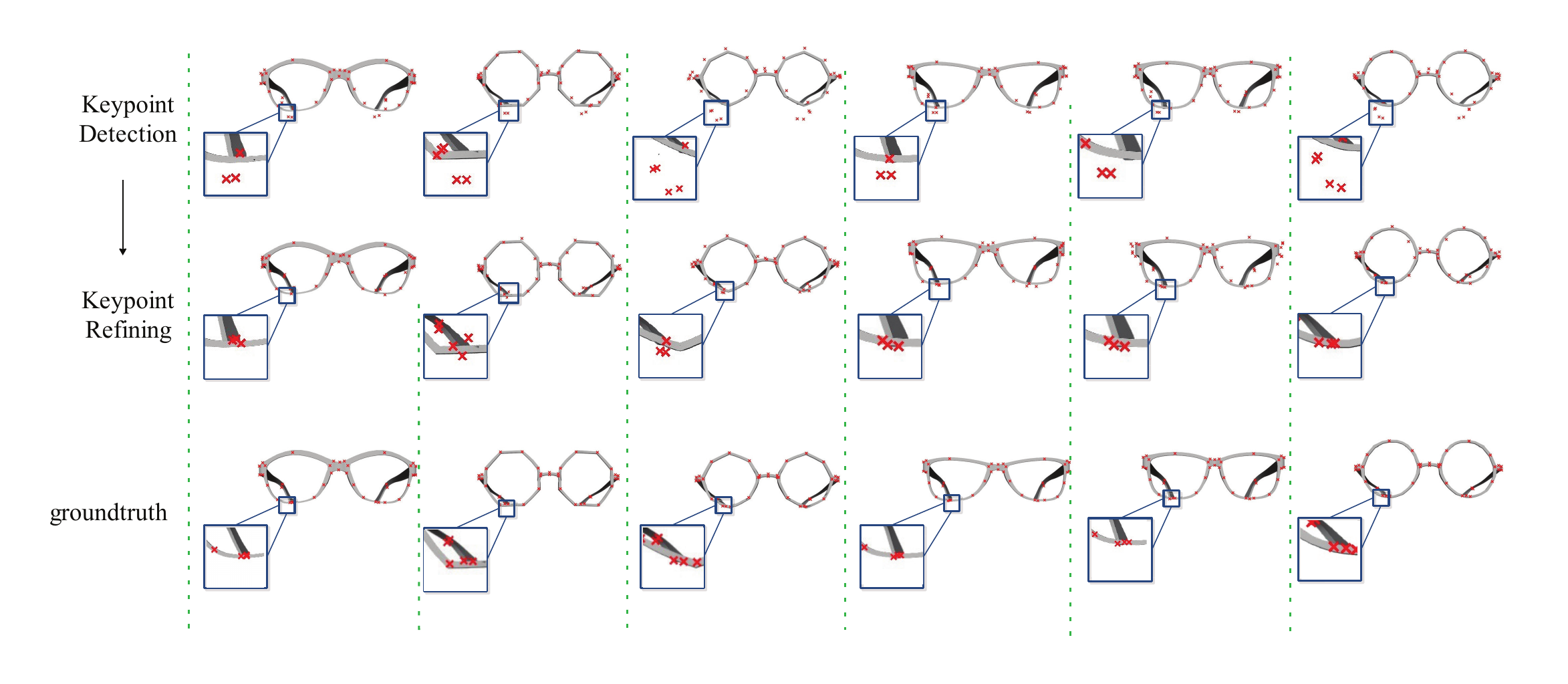}
	\caption{Examples of the keypoint detection and refining on different kinds of eyeglasses frames. The first row shows initially detected keypoints by the proposed keypoint detector only. The second row gives the refined keypoints by the proposed keypoint refiner. The last row illustrates the groundtruth keypoints.}
	\label{Fig:keypoint_detection}
\end{figure*}

\begin{figure}[!t]
	\centering
	\includegraphics[width=0.3\textwidth]{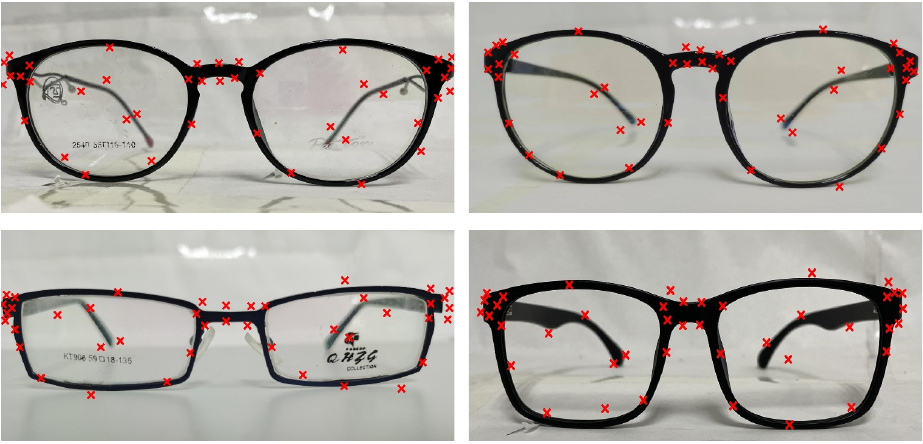}
	\caption{Keypoint detection and refining results on 4 real eyeglasses frames.}
	\label{Fig:keypoint_detection_real}
\end{figure}

\subsection{Keypoint Detection and Refining}
\label{Sec:Keypoint Detection}

To evaluate the performance of the keypoint detection and refining, we use the regressed coordinates of detected 2D keypoints by the keypoint detector and refiner for all images in the testing dataset. 
A keypoint detection error $err$ is then computed for each eyeglasses frame, which is measured with the distance between the predicted keypoints and their ground-truth,
\begin{flalign}
&&
	err = \frac{1}{169 \times 42}\sum_{i=1}^{169}\sum_{j=1}^{42} \left \|{p^{r}_{ij}-p^{g}_{ij}} \right \|_{2} ,
 &&
	\label{eq_kperr}
\end{flalign}
where $p^{r}_{ij}$ is the $j$-th detected and refined 2D keypoint of the $i$-th image, and $p^{g}_{ij}$ is the ground-truth coordinate of the corresponding  $j$-th 2D keypoint of the $i$-th image.
Both $p^{r}_{ij}$ and $p^{g}_{ij}$ are normalized to $\left [  0, 1\right ] $ by the image size.
If the $err$ exceeds 5\%, the predicted keypoint will be considered as a failed detection. 
The PCK@5\% is also used to represent the percentage of correctly detected keypoints within a 5\% error. 

Table \ref{Tab:RefinerError} shows the average detection error and the PCK@5 of all testing images of each kind of eyeglasses frames. 
Fig. \ref{Fig:keypoint_detection} gives some examples of keypoint detection and refining of synthetic views, respectively.
Fig. \ref{Fig:keypoint_detection_real} also shows some results of keypoint detection and refining on 4 real eyeglasses frames.

\begin{table}[!t]
	\centering 
	\caption{Accuracy statistics of detected and refined keypoints by the keypoint detector and refiner module.}
	\begin{tabular}{ccc}
		\hline    
		Eyeglasses  &	Average Error (\%)	&	PCK@5 (\%) \\
		\hline
		Rectangle\_1	& 2.11 & 98.59	\\
		Rectangle\_2	& 1.73 & 99.11 \\
		Rectangle\_3	& 2.03 & 98.41 \\
		Circle		& 1.93 & 97.70 \\
		Octagonal\_1	& 2.92 & 94.51 \\
		Octagonal\_2	& 1.91 & 98.06 \\ \hline
		Average       & 2.11 & 97.73 \\
		\hline
	\end{tabular}
	\label{Tab:RefinerError}
\end{table}

\subsection{Reconstruction Evaluations}
\label{Sec:Reconstruction Evaluations}
In this section, we demonstrate the effectiveness of our method through both quantitative and qualitative experiments. 
It is worth noting that we conducted quantitative experiments only on synthetic datasets due to the difficulty of obtaining accurate measurements of real eyeglass frames. 
We performed qualitative experiments on both synthetic and real datasets to show the effectiveness of our method and its generalization capability to real-world scenarios.

To quantitatively assess the reconstruction result, we first reconstruct 3D models of all the testing images of each kind of eyeglasses frame in the testing dataset.
The reconstruction error (RE) is then used to calculate the distance between the reconstructed model and its ground truth in 3D space,
\begin{flalign}
&&
	\mathrm{RE} = \frac{1}{n_{v}}\sum_{i=1}^{{n_{v}}} || v_{i}-\tilde{v}_{i}  ||,
	\label{REerr}
 &&
\end{flalign}
where $n_{v}$ is the number of vertices of the model , $v_{i}$ is the ground-truth vertex,  and $\tilde{v}_{i}$  is the corresponding reconstructed vertex.
Note that the reconstructed model and its ground truth are normalized by the diagonal distance of the ground-truth model's bounding box.
In addition, the projection error, which is determined by the IoU (Intersection over Union), is also introduced to measure the silhouette difference between the projection of the reconstructed model and its input image.

\begin{figure*}
	\includegraphics[width=1.0\textwidth]{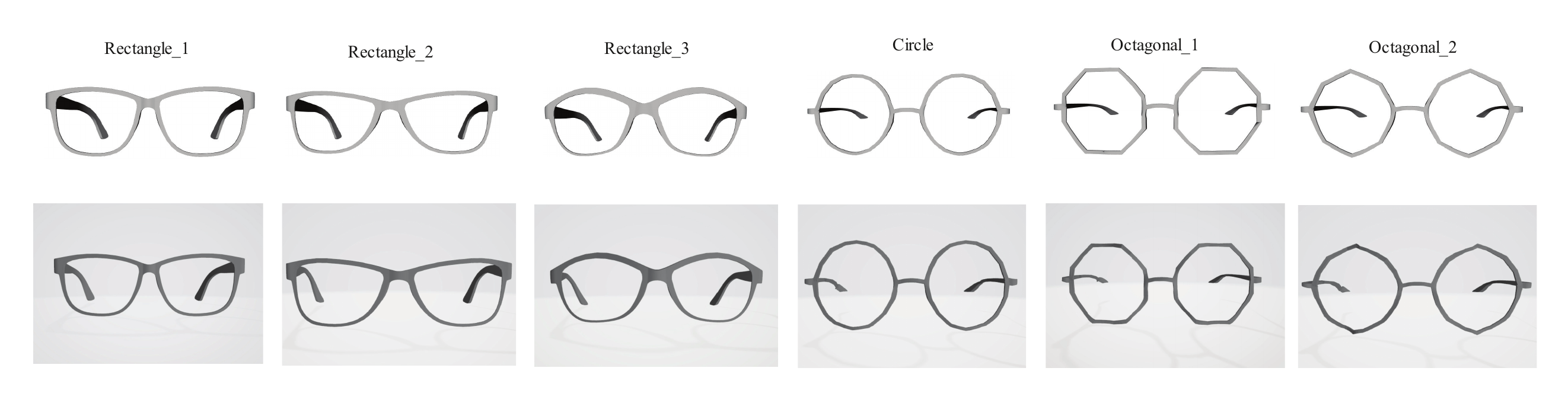}
	\caption{Reconstruction results of six eyeglasses frames selected from the testing dataset. The first row shows the input synthetic images. The second row shows corresponding reconstruction results, which are rendered by the 3D viewer software.}
	\label{Fig:reconstrution_results}
\end{figure*}

Tab. \ref{Tab:reconstructionError} reports the statistics of reconstruction and projection errors of each kind of eyeglasses frames in the testing dataset. 
Besides, Fig. \ref{Fig:reconstrution_results} shows the reconstructed models of six typical kinds of eyeglasses frames selected from the testing dataset.
It can be observed that the proposed method can faithfully recover 3D shapes of most kinds of eyeglasses frames.
However, as the shape difference between the template and the reconstructed model grows, the reconstruction error grows larger, e.g., the octagonal frame. 

\begin{table}[!t]
	\centering  
	\caption{Statistics of reconstruction error (RE) and intersection over Union (IoU). }
	\begin{tabular}{ccc}
	\hline
		Eyeglasses  &	RE	& IoU \\
		\hline
		Rectangle\_1	& 0.0750 & 0.9488 \\
		Rectangle\_2	& 0.0946 & 0.9277 \\
		Rectangle\_3	& 0.1339 & 0.9598 \\
		Circle		    & 0.0909 & 0.9315 \\
		Octagonal\_1	& 0.1192 & 0.9153 \\
		Octagonal\_2	& 0.0960 & 0.8819 \\
		\hline
		Average 	& 0.1016 & 0.9275 \\
		\hline
	\end{tabular}
	\label{Tab:reconstructionError}
\end{table}

\begin{figure*}[t]
    \centering
	\includegraphics[width=1\textwidth]{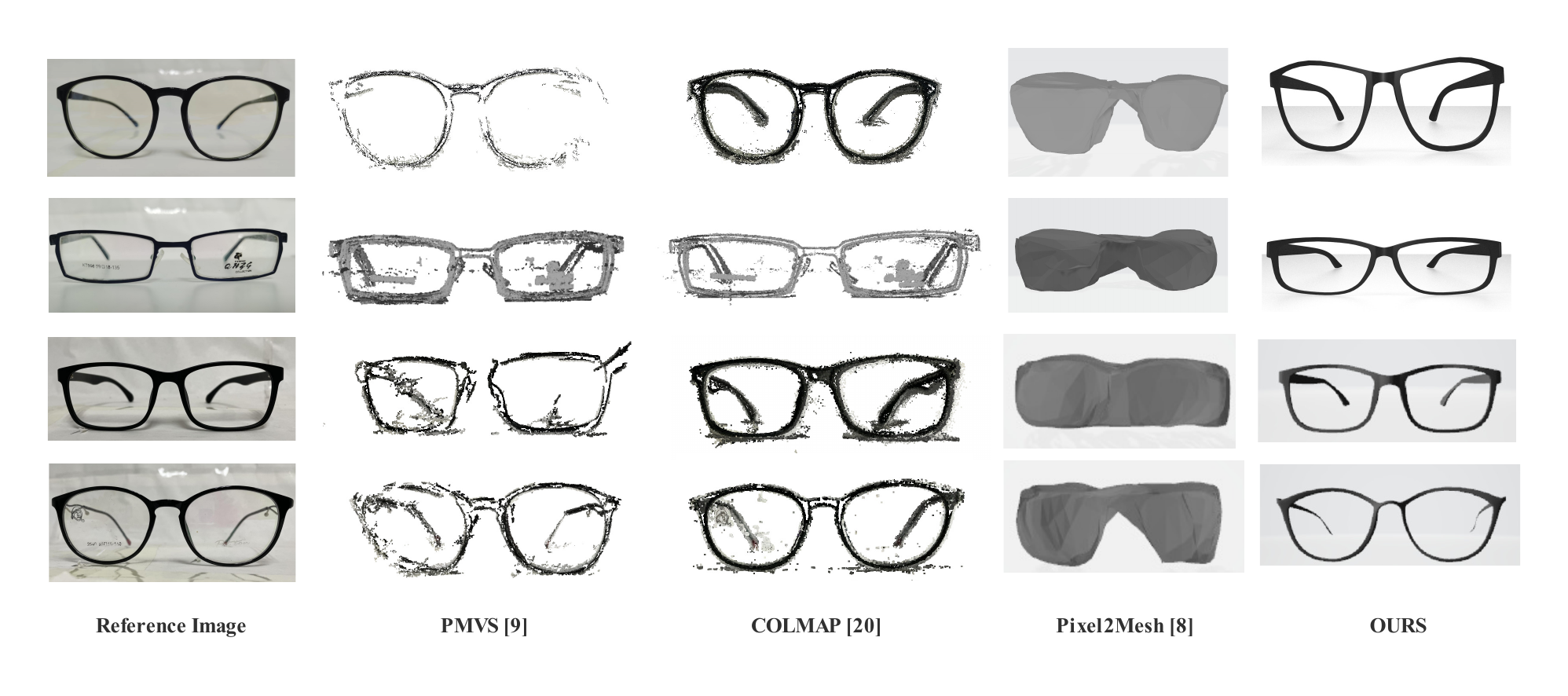}
	\caption{Reconstruction comparisons with PMVS \cite{PMVS}, COLMAP \cite{COLMAP}, and Pix2Mesh \cite{Pix2Mesh}.  }
	\label{Fig:method_comparison}
\end{figure*}

We also conduct reconstruction experiments on images of several real eyeglasses frames taken in a controlled environment. 
Fig. \ref{Fig:MethodComparisonIntro} and Fig. \ref{Fig:method_comparison} show examples of reconstruction models from five real images taken from the front, respectively.
It can be observed that our method can achieve comparable and competitive reconstruction results from real images.

\begin{figure}[!t]
	\centering
	\includegraphics[width=0.5\textwidth]{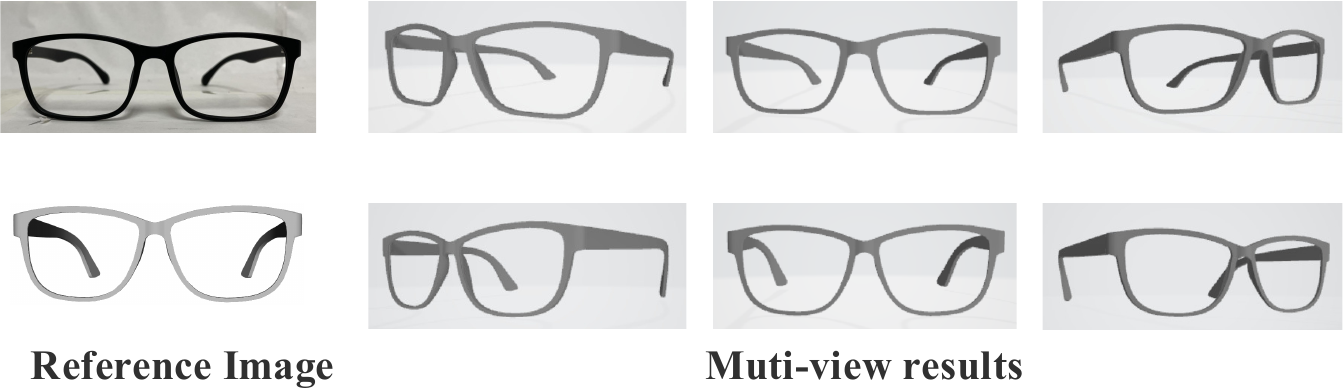}
	\caption{Reconstruction results of different views.}
	\label{Fig:muti_view_results}
\end{figure}

To comprehensively show the effect of our reconstruction method, we display the multi-view reconstruction results of two different frames, one from the synthetic dataset and one from a real image input, as illustrated in Fig. \ref{Fig:muti_view_results}.

\subsection{Comparisons with State-of-the-Art Methods}

In this paper, to demonstrate the effectiveness, we compare our method with PMVS \cite{PMVS}, COLMAP \cite{COLMAP}, and Pix2Mesh \cite{Pix2Mesh}.
PMVS \cite{PMVS} and  COLMAP \cite{COLMAP} are multi-view stereo (MVS) algorithms while the Pix2Mesh \cite{Pix2Mesh} is a deformation-based single-view reconstruction algorithm.
There are 68 input images for PMVS \cite{PMVS} and  COLMAP \cite{COLMAP} .
As shown in Fig. \ref{Fig:method_comparison}, the reconstruction results given by multi-view stereo algorithms contain significant noise and missing parts, let alone the required multiple input images and post-processing.
Among the two deformation-based methods, the Pix2Mesh can only recover the basic shape of the frame while lacking sufficient reconstruction details.
By leveraging the prior knowledge of eyeglass frames, such as symmetry, topology, etc., our method can restore the exact shape of the frame while also preserving the grid structure.

It should be pointed out that, to the best of our knowledge, there are no reconstruction algorithms specially designed for thin eyeglasses frames in the literature.
And, as discussed in Section \ref{subsec:thinReconstruction}, existing reconstruction methods for thin structures are all designed for specific categories.
They exploit prior knowledge of specific objects with thin structures for the reconstruction, which can not be used for the eyeglasses frame reconstruction.
Therefore, we do not compare our method with these methods in this paper.

\subsection{Ablation Study}

\begin{table}[!t]
\centering
	\caption{Accuracy statistics of the keypoint detection with different backbones.}
	\begin{tabular}{ccc}
		\hline
		Backbone  &	Average Error(\%)	&	PCK@5(\%)	 \\
		\hline
		DenseNet	&	2.56	&	87.89	\\
		ResNet152		&	7.38	&	28.56	\\
		VGG19			&	9.38	&	26.13	\\
		Ours &	2.11	&	97.73 \\
	 \hline
	\end{tabular}
	\label{Tab:BackBoneComparison}
\end{table}

\subsubsection{Keypoint Detector}

To evaluate the effect of our keypoint detector network, the VGG\cite{2015VGG}, ResNet\cite{resnet}, and DenseNet\cite{densenet} are employed as the feature extraction network, respectively. 
To be fair, we make some minor changes to their network structures to ensure that the input image size is $1024 \times 1024$. 
Tab. \ref{Tab:BackBoneComparison} shows the keypoint detection accuracy using different backbones.
This demonstrates that the dense connection between convolution layers can keep more information for downstream tasks while also allowing the gradient to easily transfer through the network. 

\begin{table}[!t]
	\caption{Accuracy metrics of the keypoint detection without and with the keypoint refiner. }
	\centering 
 \scalebox{0.9}{
	\begin{tabular}{lcccc}
 \toprule
\multirow{2}{*}{Category} & \multicolumn{2}{c}{Average Error(\%)} & \multicolumn{2}{c}{PCK@5(\%)}\\
\cmidrule(r){2-3} \cmidrule(r){4-5} 
&  w/o refiner     &  w refiner
&  w/o refiner     &  w refiner   \\   
\midrule
Rectangle\_1	& 3.28 & 2.11 & 81.96 & 98.59\\ 	
Rectangle\_2	& 2.52  & 1.73& 86.49 & 99.11\\
Rectangle\_3	& 3.28  &2.03 & 81.95 & 98.41 \\
Circle		& 1.52  & 1.93 & 97.63& 97.70\\
Octagonal\_1	& 1.76  & 2.92 & 96.38& 94.51 \\
Octagonal\_2	& 3.06  & 1.91 & 81.16& 98.06\\
Average		& 2.56  & 2.11 & 87.59& 97.73\\
\bottomrule
\end{tabular}}
\label{Tab:KpError2}
\end{table}

\subsubsection{Keypoint Refiner}

To evaluate the effect of the keypoint refiner, we employ the mean square error to report the keypoint detection error when only the keypoint detector is used. 
The keypoint positions regressed by the keypoint detector are also normalized to $\left[ 0,1\right]$ by the image resolution. 
Table \ref{Tab:KpError2} shows the keypoint detection error without the use of the keypoint refiner and with the use of the it. 
From Table \ref{Tab:KpError2}, it can be seen that the proposed keypoint refiner can improve the precision of keypoint estimation by 0.45\% and increase PCK@5 by 9.84\%, respectively.

\subsubsection{Loss Terms}

We perform ablation studies to examine the contribution of each loss function quantitatively and qualitatively in this section.
\begin{figure}
	\includegraphics[width=0.5\textwidth]{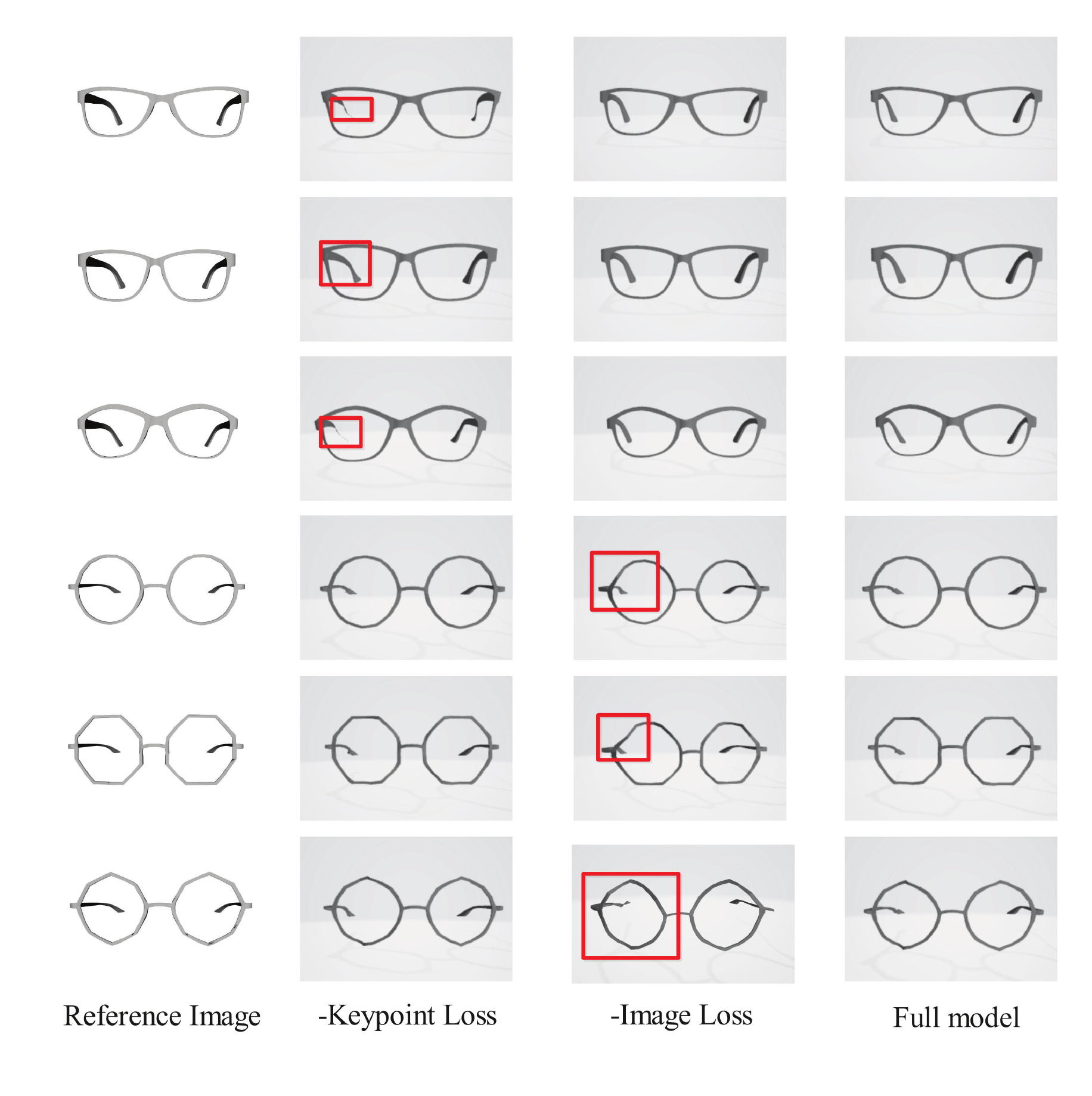}
	\caption{Ablation studies on keypoint loss and image loss. The columns from left to right show the reconstruction results without keypoint loss, without image loss, and with all loss functions, respectively.}
	\label{Fig:ablation_kpimg}
\end{figure}

\textbf{Keypoint Loss}.
From Fig. \ref{Fig:ablation_kpimg}, we can see that without the keypoint loss constraint, the reconstructed temples of the rectangle eyeglass frames are very thin and unrealistic.

\textbf{Image Loss}.
As shown in Fig. \ref{Fig:ablation_kpimg}, the reconstruction process of the octagonal eyeglass frames lacks the image loss constraint, so even though the predefined keypoints can be adjusted to the desired positions, the reconstruction results do not match the input images and are somewhat distorted. 
\begin{figure}
\centering	\includegraphics[width=0.5\textwidth]{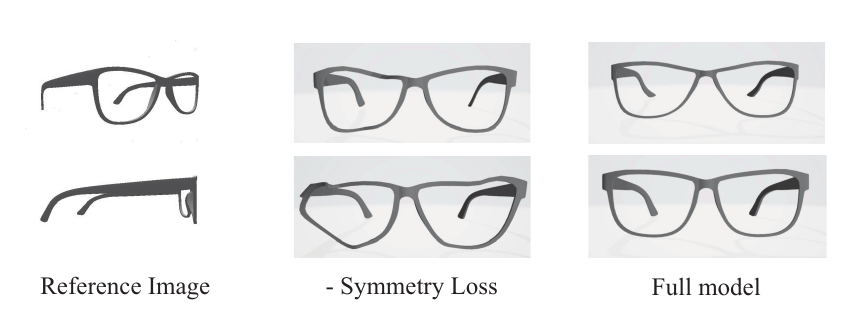}
	\caption{Ablation studies on symmetry loss. The columns from left to right show the input image, the reconstruction result without symmetry loss, and the reconstruction result with symmetry loss, respectively.}
	\label{Fig:ablation_sym}
\end{figure}

\textbf{Symmetry Loss}.
It can be seen from Fig. \ref{Fig:ablation_sym} that the symmetry loss function will make our reconstruction result more symmetrical and reasonable when the input image is a side view.

\textbf{Silhouette Loss}.
We can see from Table \ref{Tab:ablation studu on iou and laplacian loss} that 
the IoU metrics of the six eyeglass frames decrease when the silhouette loss is removed from all loss functions. 
And, the similarity with the original image is reduced. 

\textbf{Smooth regularization}.
As can be seen from Table \ref{Tab:ablation studu on iou and laplacian loss}, the projection error is reduced when the smooth regularization is removed from all loss functions. This means that the Laplacian loss can make our free-form deformation process smoother and faster, and achieve better results with the same number of iterations.

\begin{table}[!t]
\centering
	\caption{IoU of the reconstruction result from the front view. The columns from left to right show the results without silhouette loss, without smooth regularization, and with all losses, respectively.}
 \scalebox{0.8}{
	\begin{tabular}{lccc}
		\hline
		Eyeglasses  & -Silhouette Loss	&	-Smooth regularization & Full model	 \\
		\hline
	    Rectangle\_1	 & 0.9482	& 0.9480 & 0.9488		\\
	    Rectangle\_2  & 0.9270	  &	0.9249 & 0.9277   \\
        Rectangle\_3  & 0.9597	& 0.9597 & 0.9598 	 \\
        Circle    & 0.9313	&	0.9314 &  0.9316     \\
        Octagonal\_1  & 0.9136  &  0.9123  & 0.9153  \\
        Octagonal\_2  & 0.8810   & 0.8817  &  0.8820  \\
	 \hline
	\end{tabular}}
	\label{Tab:ablation studu on iou and laplacian loss}
\end{table}

\subsection{Limitations}

\begin{figure}[!t]
	\includegraphics[width=0.5\textwidth]{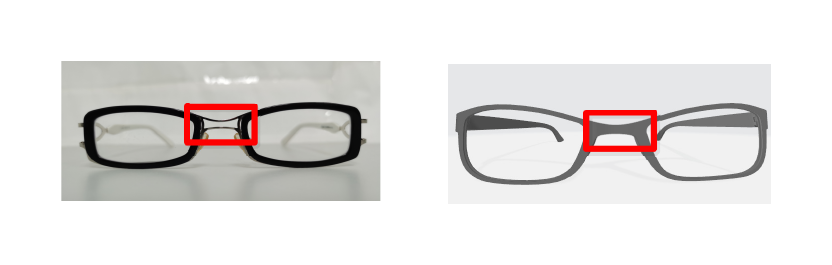}
	\caption{Reconstruction result of an eyeglasses frame with the different topology from the template. }
	\label{Fig:Limitation-structures}
\end{figure}

First, essentially, our method is a kind of deformation-based reconstruction technique that is restricted to the topology of the template.
As a result, it can only reconstruct surfaces with fixed topology. 
Fig. \ref{Fig:Limitation-structures} gives an example of an eyeglasses frame with the bridge topology that differs from the template.
It can be observed that our method cannot recover the structure of the bridge well.
In addition, our method is not good at the reconstruction of structures with abrupt changes in thickness or hollows.
Therefore, as a deformation-based reconstruction technique, we recommend using common full-frame eyeglasses for reconstruction.

\begin{figure}[!t]
	\includegraphics[width=0.5\textwidth]{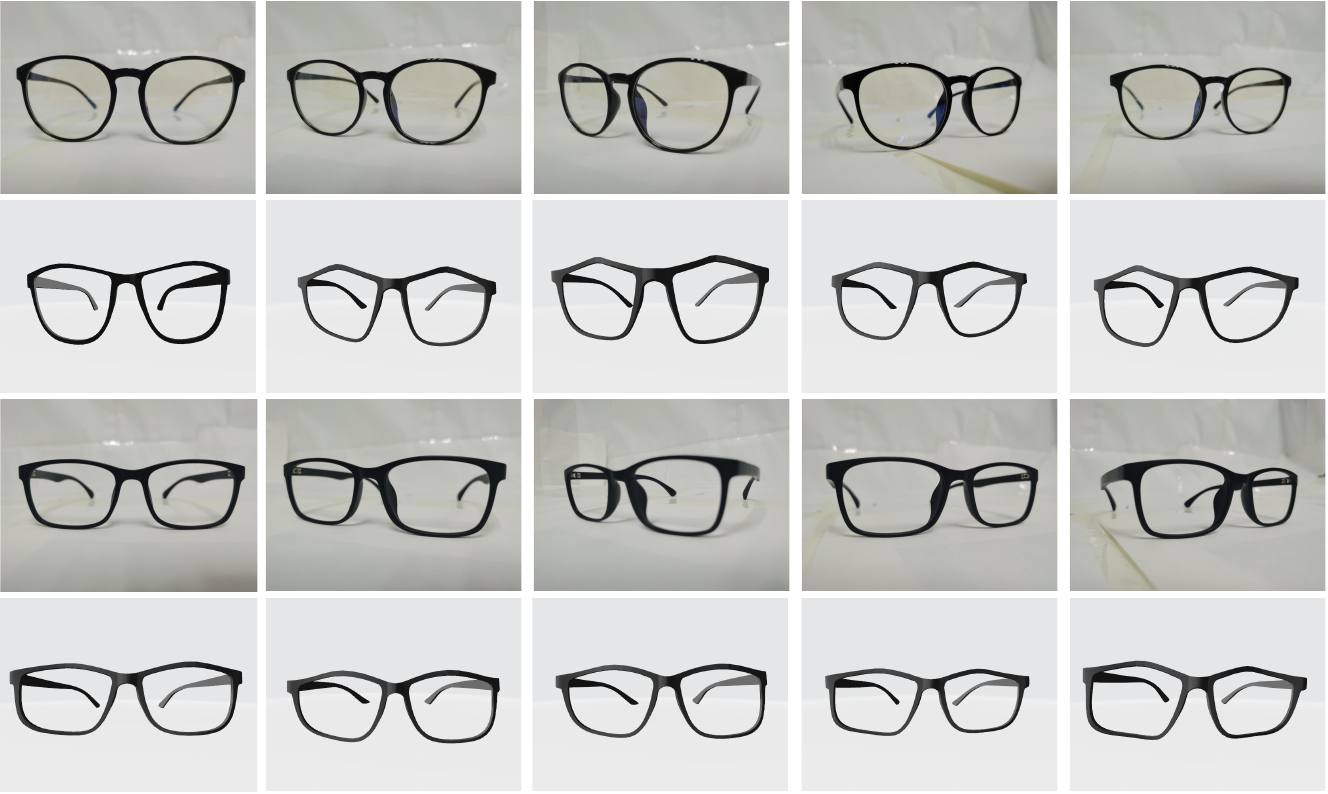}
	\caption{Reconstruction results from different viewpoints. }
	\label{Fig:Limitation-view}
\end{figure}

Second, our method requires that the input image be captured from the front as much as possible for easy and accurate keypoint detection and reconstruction, as illustrated in Fig. \ref{Fig:method_comparison}. 
Fig. \ref{Fig:Limitation-view} shows some reconstruction results of same eyeglasses frames from different viewpoints.
From Fig. \ref{Fig:Limitation-view}, we can see that, as the shooting angle increases, the reconstruction accuracy begins to decrease gradually, especially when the occluded parts of the frame increase.
This is because the large shooting angle may lead to more inaccurate detected keypoints and estimated camera pose, resulting in unfaithful reconstruction results.
Therefore, as a type of single-view reconstruction framework, it is recommended to take the input image from the front of the eyeglasses frame to obtain highly precise reconstructed models.

Additionally, we recommend selecting a background that contrasts well with the color of the eyeglass frames when capturing images. This will facilitate the processing of the input images for keypoint estimation, camera pose estimation, and reconstruction.

\section{Conclusion}

In this paper, we propose a novel deformation-based method for high quality reconstruction of 3D full-frame eyeglasses with thin structures from a single-view image. 
A coarse-to-fine strategy is proposed to accurately detect predefined keypoints to estimate the camera pose of the input image. 
Then, an unsupervised free-form deformation technique that integrates with differentiable rendering is designed to progressively deform the template mesh to output the final reconstructed model.
In the future, we would like to reconstruct eyeglasses frames from images taken in the wild, providing users with a better experience. 
Another extension is to consider the self-supervised reconstruction method to reduce reliance on datasets. 








\appendix

\section{Details of Keypoint Detector}

The keypoint detector consists of an encoder and a regressor, as illustrated in Fig.\ref{Fig:KeypointDetector}. 

The encoder, which consists of 8 dense blocks with a growth rate of 2, employs dense connections to get high-level features from an input RGB image with a resolution of $1024 \times 1024$.
The input RGB image is first converted to grey scale one.
The definition of growth rate can be found in \cite{densenet}.
A convolution with a kernel size of $7 \times 7$ and four output channels is first performed on the input image.
Each dense block contains several convolution blocks, each of which is composed of two convolution layers with kernel sizes of $1 \times 1$ and $3 \times 3$, respectively.
Each convolution block has dense connections to its following convolution blocks in the same dense block.
To concatenate feature maps of different sizes, a transition layer is inserted between two dense blocks, which consist of a batch normalization layer and a $1 \times 1$ convolutional layer followed by a $2 \times 2$ average pooling layer.
The encoder produces $1376 \times 4^{2}$ feature maps in total.

The regressor consists of five fully-connected (FC) layers.
The first layer, which has 22016 channels, expands the feature maps into a one-dimensional vector. 
The channel number of the following FC layer is reduced by half for every FC layer.
The final layer performs 42-way keypoint regression and thus contains 84 channels.
Note that all hidden layers are equipped with the rectification (ReLU) non-linearity \cite{AlexNet}.

\section{Details of Keypoint Refiner}

The keypoint refiner consists of three parts: an encoder, a maxpooling layer, and a decoder, as illustrated in Fig. \ref{Fig:Refiner}.

The encoder, which consists of four one-dimensional convolution layers, applies feature transformations to the input. 
The first convolution layer filters the input with 16 kernels of size $1 \times 7$ to extract global features.
The final three layers are one-dimensional convolution layers with 64, 256, and 1024 kernels of a smaller size $1 \times 3$ to extract local features, respectively. 
Each convolution layer is followed by a ReLU layer to perform non-linearity on the feature transformation.

The maxpooling layer is used to keep the maximum value for each channel and thus transform the features into a vector.  
We train the refiner so that the encoder and the maxpooling layer can learn to use the vector to represent low-frequency features of keypoints that have been found.

The decoder converts the vector into refined positions of keypoints $p^{r}$ with a set of 1D transposed convolution layers.


\bibliographystyle{unsrt}

\bibliography{cas-refs}



\end{document}